\documentclass[runningheads]{llncs}
\usepackage{graphicx}
\usepackage{siunitx}
\usepackage{amsmath,amssymb} 
\usepackage{color}
\usepackage{multirow}
\usepackage{hhline}
\usepackage[pscoord]{eso-pic}
\usepackage{subcaption}
\newcommand{\placetextbox}[3]{
  \setbox0=\hbox{#3}
  \AddToShipoutPictureFG*{
    \put(\LenToUnit{#1\paperwidth},\LenToUnit{#2\paperheight}){\vtop{{\null}\makebox[0pt][c]{#3}}}%
  }%
}%

\begin{document}

\placetextbox{0.35}{0.13}{Published in ECCV 2018 Workshops}
\placetextbox{0.375}{0.115}{\textcopyright  Springer, LNCS 11129, pp. 473-487, 2019 }
\placetextbox{0.42}{0.10}{The final authenticated publication is available online at}
\placetextbox{0.38}{0.085}{ https://doi.org/10.1007/978-3-030-11009-3\_29}%

%
\title{Paired 3D Model Generation with Conditional Generative Adversarial Networks} 

\titlerunning{Paired 3D Model Generation with CGAN}
%
\author{Cihan \"Ong\"un \and
Alptekin Temizel} 

%
\authorrunning{C. \"Ong\"un and A. Temizel}
%

\institute{ Middle East Technical University, Ankara, Turkey 
\linebreak
\email{\{congun,atemizel\}@metu.edu.tr}}
\maketitle              
\begin{abstract}
Generative Adversarial Networks (GANs) are shown to be successful at generating new and realistic samples including 3D object models. 
Conditional GAN, a variant of GANs, allows generating samples in given conditions. However, objects generated for each condition are different 
and it does not allow generation of the same object in different conditions. In this paper, we first adapt conditional GAN, which is originally 
designed for 2D image generation, to the problem of generating 3D models in different rotations. We then propose a new approach to guide 
the network to generate the same 3D sample in different and controllable rotation angles (sample pairs). Unlike previous studies, the proposed 
method does not require modification of the standard conditional GAN architecture and it can be integrated into the training step of any 
conditional GAN. Experimental results and visual comparison of 3D models show that the proposed method is successful at generating model pairs in 
different conditions.

\end{abstract}
\section{Introduction}
\label{sect:intro}

While 3D technology mostly focuses on providing better tools for humans to scan, create, modify and visualize 3D data, recently there has been 
an interest in automated generation of new 3D object models. Scanning a real object is the most convenient way to generate digital 3D object
 models, however, this requires availability of real-life objects and each of these objects needs to be scanned individually. More crucially, 
 it does not allow creating a novel object model. Creating a novel object model is a time consuming task requiring human imagination, effort and 
 specialist skills. So it is desirable to have an automated system facilitating streamlined generation of 3D object content. 

Generative models have recently become mainstream with their applications in various domains. Generative Adversarial Networks (GANs) 
\cite{goodfellow2014generative} have been a recent breakthrough in the field of generative models. GANs provide a generic solution for various 
types of data leveraging the power of artificial neural networks, particularly Convolutional Neural Networks (CNN). On the other hand, use of 
GANs brings out several challenges. While stability is the most fundamental problem in GAN architecture, there are also domain specific challenges. 

Standard GAN model generates novel samples from an input distribution. However, the generated samples are random and there is no control over them as the input noise and the desired features are entangled. While some solutions attack the entanglement problem \cite{chen2016infogan}, some propose new types of GANs for specific purposes. Conditional GAN \cite{mirza2014conditional} allows controlling the characteristics of the generated samples using a condition. While these conditions could be specified, the generated samples are random and it fails to generate pair samples in different conditions \cite{liu2016coupled,mao2017aligngan}. Keeping the input value the same while changing the condition value does not generate the same output in different conditions because of the entanglement problem. The representation between input and output sample is entangled in such a way that changing condition value changes the output completely. There are many studies for learning joint distributions to generate novel pair samples. Most of them uses modified GAN architectures, complex models or paired training data as described in the Related Works section.

In this study, we propose a new approach to generate paired 3D models with Conditional GANs. Our method is integrated as an additional training step to Conditional GAN without changing the original architecture. This generic solution provides flexibility such that it is applicable to any conditional GAN architecture as long as there is a metric to measure the similarity of samples in different conditions. Also the system can be trained with paired samples, unpaired samples and without any tuples of corresponding samples in different domains.
	 
In section 2, we describe the GAN architecture and the related works. The proposed method is given in section 3 and experimental evaluation and results are provided in section 4. Conclusions and future work are given in section 5.

\section{Related Works}
\label{sect:relatedworks} 
\begin{figure}
\centering
\begin{subfigure}{.7\textwidth}
  \centering
  \includegraphics[width=\textwidth]{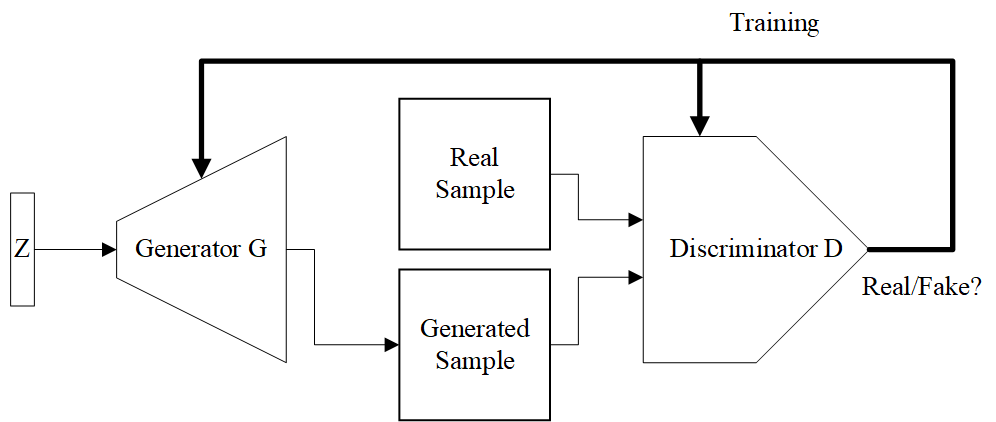}
  \caption{}
  \label{fig:sfig1}
\end{subfigure}%

\begin{subfigure}{.7\textwidth}
  \centering
  \includegraphics[width=\textwidth]{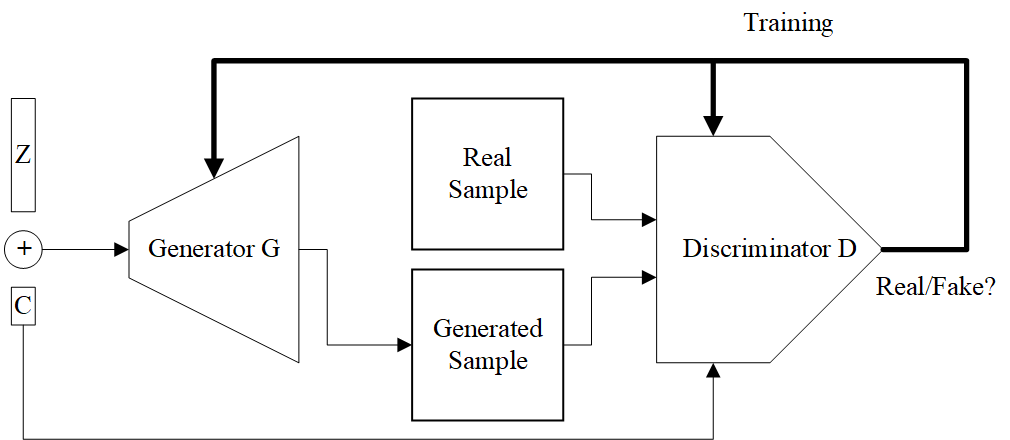}
  \caption{}
  \label{fig:sfig2}
\end{subfigure}
\caption{(a) Standard GAN and (b) conditional GAN architectures.}
\label{fig:fig1}
\end{figure}

GAN architecture (Fig. \ref{fig:fig1}(a)) consists of a generator model $\mathnormal{G}$ and a discriminator model $\mathnormal{D}$ \cite{goodfellow2014generative}. Generator model takes an input code and generates new samples. Discriminator model takes real and generated samples and tries to distinguish real ones from generated ones. Generator and discriminator are trained simultaneously so that while generator learns to generate better samples, discriminator becomes better at distinguishing samples resulting in an improved sample generation performance at the end of the training.

	If GAN is trained with training data $\mathbf{x}$ for discriminator $D$ and sampled noise z for generator $G$, $D$ is used to maximize the correctly labeled real samples as real $\log(D(x))$ and generated samples as fake $\log(1 - D(G(z)))$. On the other hand, generator $G$ tries to fool the discriminator to label the generated data as real so $G$ is used for minimizing $\log(1 - D(G(z)))$ . These two models duel each other in a min-max game with the value function $V(D,G)$. The objective of the whole system can be formulated as: 

\begin{equation}
\begin{split}
min_{_{G}}max_{_{D}}V(D,G) &= E_{x\sim p_{data}(x)}[log~D(x)]\\
&+  E_{z\sim p_{z}(z)}[log~ (1-D(G(z)))].
\end{split}
\end{equation}

Use of CNN based GANs \cite{radford2015unsupervised} is popular in 2-D image domain with various applications.  Pix2pix \cite{isola2017image} is a general-purpose GAN based solution to image-to-image translation problems and it has been shown to be effective at problems such as synthesizing photos from label maps, reconstructing objects from edge maps, and colorizing images. It uses GANs in conditional settings for image-to-image translation tasks, where a condition is given on an input image to generate a corresponding output image. Another application of GAN is style transfer \cite{luan2017deep}. For an input image, the system can transfer the style of the reference image including time of the day, weather, season and artistic edit to the target. Perceptual Adversarial Network (PAN) \cite{wang2018perceptual} provides a generic framework for image-to-image transformation tasks such as removing rain streaks from an image (image de-raining), mapping object edges to the corresponding image, mapping semantic labels to a scene image and image inpainting.

The fundamental principle of GANs, i.e. using two different models trained together, causes stability problems. These two models must be in equilibrium to work together in harmony. Since the architecture is based on dueling networks, during the training phase, one of the models could overpower the other, causing a stability problem. Wasserstein GAN \cite{arjovsky2017wasserstein} proposes a new distance metric to calculate the discriminator loss where Wasserstein distance (Earth-Mover distance) is used to improve the stability of learning and provide useful learning curves. In \cite{che2016mode} several approaches are introduced for regularizing the system to stabilize the training of GAN models.

Generating 3D models with GANs is a relatively new area with a limited number of studies. The first and the most popular study uses an all-convolutional neural network to generate 3D objects \cite{wu2016learning}. In this work, the discriminator mostly mirrors the generator and 64x64x64 voxels are used to represent 3D models. Wasserstein distance \cite{smith2017improved} is employed by normalizing with gradient penalization as a training objective to improve multiclass 3D generation. In another study an autoencoder network is used to generate 3D representations in latent space \cite{achlioptas2017representation}. GAN model generates new samples in this latent space and these samples are decoded using the same autoencoder network to obtain 3D point cloud samples. 3D meshes can also be used to train a GAN \cite{jiang2017hierarchical} to produce mesh-based 3D output. To overcome the difficulty of working with mesh data, input data is converted to signed distance field, then processed with two GAN architectures: low-frequency and high-frequency generator. After generating high and low-frequency samples, outputs are combined to generate a 3D mesh object.

Generator model of GAN uses a simple input noise vector $\mathbf{z}$ and it is possible that the noise will be used by the generator in a highly entangled way, causing the input vector $\mathbf{z}$ not correspond to semantic features of the output data.  InfoGAN \cite{chen2016infogan} is a method proposed to solve entanglement problem. To make a semantic connection between input noise vector and output data, a simple modification is presented to the generative adversarial network objective that encourages it to learn interpretable and meaningful representations. Generator network is provided with both the incompressible noise $\mathbf{z}$ and the latent code $\mathbf{c}$, so the form of input data becomes $(z, c)$. After necessary optimizations for combining these values, expected outputs can be generated with given parameters.

Conditional GAN (Fig. \ref{fig:fig1} (b)) is an extended version of GAN \cite{mirza2014conditional} conditioning both generator and discriminator on some extra information. While standard GAN models generate samples from random classes, CGANs can generate samples with a predetermined class for any input distribution such as generating specific digits by using class labels as condition in MNIST dataset.

Input noise $\mathbf{z}$ and condition value $\mathbf{y}$  are concatenated to use as input to the generator $G$. Training data $\mathbf{x}$ and condition value $\mathbf{y}$ are concatenated to use as input to the discriminator $D$. With this modification, the objective function of conditional GAN can be formulated as follows:

\begin{equation}
\begin{split}
min_{_{G}}max_{_{D}}V(D,G) &= E_{x\sim p_{data}(x)}[log~D(x|y)] \\
&+  E_{z\sim p_{z}(z)}[log~ (1-D(G(z|y)))].
\end{split}
\end{equation}

Conditional GANs can generate samples in given conditions, however they are not able to generate pairs for the same input and different condition values. Coupled GAN (CoGAN) \cite{liu2016coupled} is a new network model for learning a joint distribution of multi-domain images. CoGAN consists of a pair of GANs, each having a generative and a discriminator model for generating samples in one domain. By sharing of weights, the system generates pairs of images sharing the same high-level abstraction while having different low-level realizations. DiscoGAN \cite{kim2017learning} aims to discover cross-domain relations with GANs. A similar approach is used in CycleGAN \cite{zhu2017unpaired} where an image is used as input instead of a noise vector and it generates a new image by translating it from one domain to another. SyncGAN \cite{chen2018syncgan} employs an additional synchronizer model for multi-modal generation like sound-image pairs. AlignGAN \cite{mao2017aligngan} adopts a 2-step training algorithm to learn the domain-specific semantics and shared label semantics via alternating optimization.

\section{Proposed Method}
\label{sect:method}
While the standard GAN model can generate realistic samples, it basically generates random samples in given input distribution and does not provide any control over these generated samples. For example, when a chair dataset is used to train the network, it generates chairs without any control over its characteristics such as its rotation. Conditional GANs provide control over the generated samples by training the system with given input conditions. For example, if rotation is used as a condition value for chair dataset, system can generate samples with a given rotation. 

	For both standard GAN and conditional GAN, the representation between the input and the output is highly entangled such that changing a value in the input vector changes the output in an unpredictable way. For example, for chair dataset, each chair generated by standard GAN would be random and it would be created in an unknown orientation. Conditional GAN allows specification of a condition Input vector $\mathbf{z}$ and condition value $\mathbf{y}$  are concatenated and given together as input to the system so input becomes $(z|y)$. As the condition value $\mathbf{y}$  is also an input value, changing the condition also changes the output. Even if the input vector $\mathbf{z}$ is kept the same, the model generates different independent samples in given conditions and does not allow generating the same sample in different conditions \cite{liu2016coupled,mao2017aligngan}. For example, for chair dataset, if the condition is rotation, system generates a chair in first rotation and a different chair in different rotation. As these objects are different, they cannot be merged at a later processing stage to create a new sample with less artifacts.
	
\begin{figure}
\centering
\includegraphics[width=.7\textwidth]{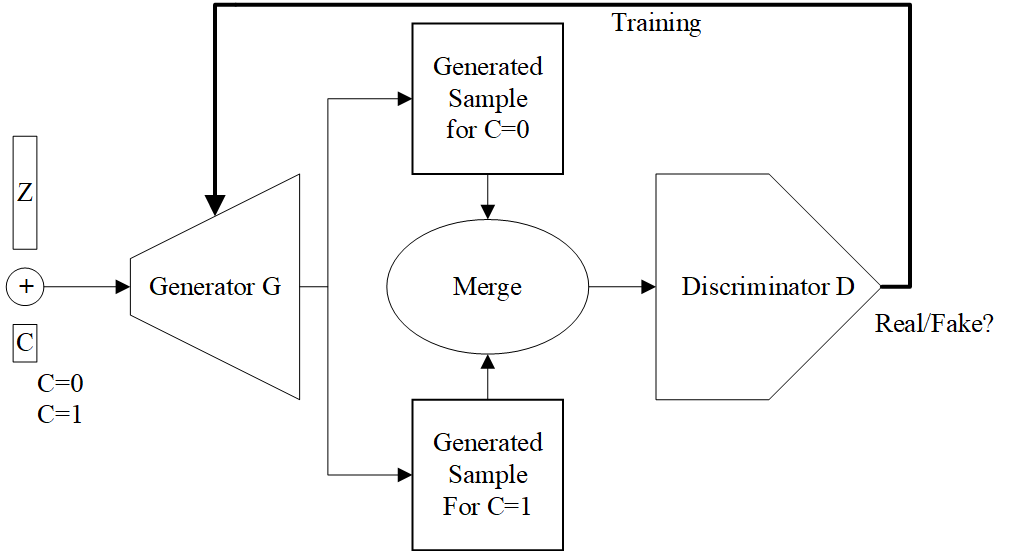}
\caption{Proposed method illustrated for 2-condition case.} \label{fig2}
\end{figure}

	To overcome this problem, we propose incorporating an additional step in training to guide the system to generate the same sample in different conditions. The pseudo code of the method is provided in Algorithm 1 and Fig. \ref{fig2} illustrates the proposed method for the 2-condition case.  We use standard conditional GAN model and training procedure to generate samples by keeping the input vector $\mathbf{z}$ the same and changing the condition value.  Generator function is defined as $G(z|y)$ for input vector $\mathbf{z}$ and condition value $\mathbf{y}$ . We can define the function for same input vector and $n$ different conditions as $G(z|y_n)$ and the domain specific merging operator as $M(G(z|y_n))$. We feed the merged result to discriminator to determine if it is realistic so the output of discriminator is $D(M(G(z|y_n)))$ . Since the proposed method is an additional step to standard conditional GAN, it is a new term for the min-max game between generator and discriminator. The formulation of proposed method can be added to standard formulation to define the system as a whole. The objective function of conditional GAN with proposed additional training step can be formulated as follows: 

\begin{equation}
\begin{split}
min_{_{G}}max_{_{D}}V(D,G) &= E_{x\sim p_{data}(x)}[log~D(x|y)] \\
&+  E_{z\sim p_{z}(z)}[log~ (1-D(G(z|y)))] \\
&+  E_{z\sim p_{z}(z)}[log~ (1-D(M(G(z|y_{n}))))].
\end{split}
\end{equation}

As expected the system generates n different samples at n different rotations even though the input vector is the same.  However as their rotations are specified by the condition, they are known. We then merge these samples to create a single object by first aligning these samples and then taking the average of the values for each voxel, similar to taking the intersection of 3D models. The merged model is then fed into the discriminator to evaluate whether it is realistic or not:

\begin{itemize}
\item If generated objects are different (as expected at the beginning), the merged model will be empty or meaningless. The discriminator will label the merged result as fake and the generator will get a negative feedback.
\item If generated objects are realistic and similar, the merged model will also be very similar to them and to a realistic chair model. The discriminator is likely to label the merged object as real and the generator gets a positive feedback.
\end{itemize}

By this additional training step, even if the generated samples are realistic, system gets negative feedback unless the samples are similar. We enforce the system to generate similar and realistic samples in different conditions for the same input vector. 

Note that the merge operation is domain specific and could be selected according to the target domain.

\vbox{
\hrule
\vspace{1mm}
\noindent \textbf{Algorithm 1.} Conditional GAN training with proposed method for n-conditions
\hrule
\vspace{1mm}
\noindent \textbf{Input:} Real samples in n conditions: $X_0,X_1,\dotsb, X_n$ input vector: $Z$,  condition values: $C_0, C_1,\dotsb, C_n$\\
Initialize network parameters for discriminator $D$, Generator $G$ and merge operation $M$\\*\\*
\textbf{for} number of training steps \textbf{do}\\*\\*
// Standard conditional GAN
\begin{itemize}
\begin{small}
\item[$\bullet$]  Update the discriminator using $X_0,X_1,\dotsb, X_n$ with $C_0, C_1,\dotsb, C_n$ respectively
\item[$\bullet$] Generate samples $S_0, S_1,\dotsb, S_n$ using vector $Z$ with $C_0, C_1,\dotsb, C_n$ respectively
\item[$\bullet$] Update the discriminator using $S_0, S_1,\dotsb, S_n$ with $C_0, C_1,\dotsb, C_n$ respectively
\item[$\bullet$] Update the generator using $S_0, S_1,\dotsb, S_n$ with $C_0, C_1,\dotsb, C_n$ respectively
\end{small}
\end{itemize}
// Proposed method
\begin{itemize}
\item[$\bullet$]Align $S_1,\dotsb, S_n$  with $S_0$
\item[$\bullet$]Merge $S_0, S_1,\dotsb, S_n: M(S_0, S_1,\dotsb, S_n)$
\item[$\bullet$]Feed merged sample to the discriminator with condition $C_0$
\item[$\bullet$]Update the generator using the discriminator output
\end{itemize}
\textbf{end for}
\vspace{1mm}
\hrule
}

\section{Experiments}
\label{sect:expert}
To test the system we used ModelNet \cite{wu20153d} dataset to generate 3D models for different object classes (e.g. chair, bed, sofa). We adapted the conditional GAN for the problem of generation of 3D objects. We then evaluated the proposed method for 2-conditional and 4-conditional cases. Visual results as well as objective comparisons are provided at the end of this section.

\subsubsection{ModelNet dataset:} This dataset contains a noise-free collection of 3D CAD models for objects. There are 2 manually aligned subsets with 10 and 40 classes of objects for deep networks. While the original models are in CAD format, there is also voxelized version \cite{wu20153d}. Voxels are basically binary 3D matrices, each matrix element determines the existence of unit cube in the respective location. Voxelized models have $30 \times 30 \times 30$ resolution. The resolution is set to $32 \times 32 \times 32$ by simply zero padding one unit on each side. For the experiments 3 object classes are used: chair, bed and sofa having 989, 615 and 780 samples respectively. Each sample has 12 orientations $O_1,O_2,\dotsb,O_{12}$ with 30 degrees of rotation between them. In the experiments with 2 orientations we use $O_1$, and $O_7$ which represent the object in opposite directions (\ang{0} and \ang{180}). Experiments with 4 orientations use $O_1$, $O_4$, $O_7$ and $O_{10}$ (\ang{0}, \ang{90}, \ang{180} and \ang{270}).

While there are more object classes in the dataset, either they do not have sufficient number of training samples for the system to converge (less than 500) or objects are highly symmetric such that different orientations come out as same models (round or rectangle objects). For different rotations, the system has been tested with paired input samples, unpaired (shuffled) samples or removing any correspondence between samples in different conditions by using one half of the dataset for one condition and the other half for other condition. The tests with different variants of input dataset show no significant change on the output. 

\subsubsection{Network structure:} We designed our architecture building on a GAN architecture for 3D object generation \cite{smith2017improved}. In this architecture, the generator network uses 4 3D transposed deconvolutional layers and a sigmoid layer at the end. Layers use ReLU activation functions and the generator takes a 200 dimensional vector as input. Output of the generator network is a $32 \times 32 \times 32$ resolution 3D matrix. Discriminator network mostly mirrors the generator with 4 3D convolutional layers with leaky ReLU activation functions and a sigmoid layer at the end. It takes a $32 \times 32 \times 32$ voxel grid as input and generates a single value between 0 and 1 as output, representing the probability of a sample being real. Both networks use batch normalization between all layers. Kernel size of convolutional filters is 4 and stride is 2.

\subsubsection{Adapting conditional GAN for generation of 3D models:} To generate 3D models on different rotations, we modified the aforementioned GAN architecture and converted it into a conditional GAN. Conditional value $\mathbf{y}$  is concatenated into $\mathbf{z}$ for generator input. For discriminator input, $\mathbf{y}$  is concatenated into real and generated samples as an additional channel. To train the discriminator, we feed objects on different rotations with corresponding condition values. To generate pairs, we change only the $\mathbf{y}$  and keep the $\mathbf{z}$ the same. 

\subsubsection{Training:} Since generating 3D models is a more difficult task than differentiating between real and generated ones, discriminator learns faster than generator and it overpowers the generator. If the learning pace is different between generator and discriminator, it causes instability in the network and it fails to generate realistic results \cite{goodfellow2014generative}. To keep the training in pace, we used a threshold for discriminator training. Discriminator is updated only if the accuracy is less than 95\% in the previous batch. The learning rates are 0.0025 for generator and 0.00005 for discriminator. ADAM \cite{kingma2015adam} is used for optimization with $\beta$ = 0.5 . System is trained using a batch size of 128. For 2 orientations, condition 0 and 1 are used for \ang{0} and \ang{180} respectively. For 4 orientations, condition 0, 1, 2 and 3 are used for \ang{0}, \ang{90}, \ang{180}, \ang{270} respectively.

\begin{figure}
\centering
\begin{subfigure}{.3\textwidth}
  \centering
  \includegraphics[width=\textwidth]{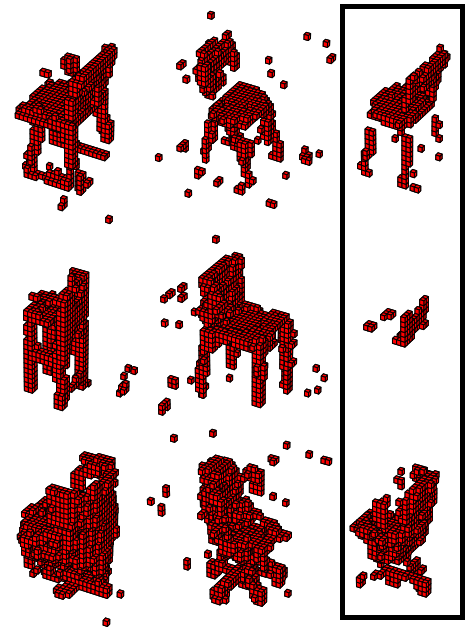}
  \caption{chair}
 \label{fig:sfig3a}
\end{subfigure}%
\qquad
\begin{subfigure}{.3\textwidth}
  \centering
  \includegraphics[width=\textwidth]{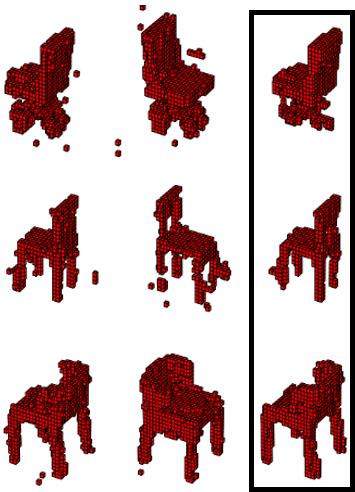}
  \caption{chair}
  \label{fig:sfig3b}
\end{subfigure}

\begin{subfigure}{.3\textwidth}
  \centering
  \includegraphics[width=\textwidth]{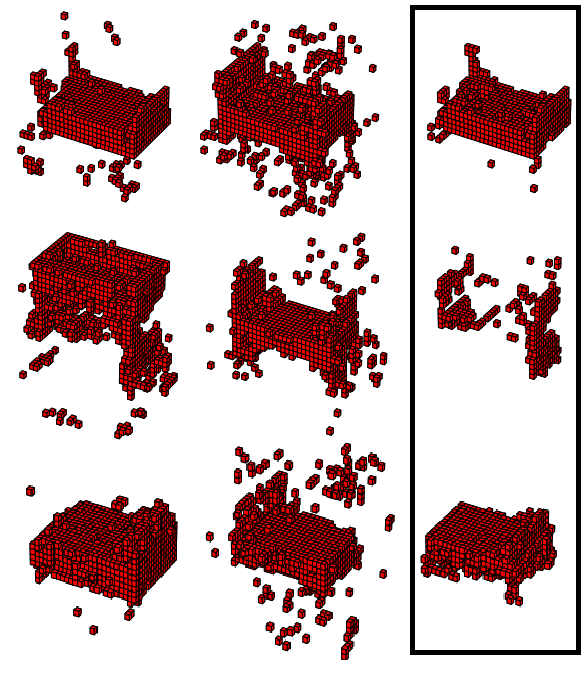}
  \caption{bed}
 \label{fig:sfig3c}
\end{subfigure}%
\qquad
\begin{subfigure}{.3\textwidth}
  \centering
  \includegraphics[width=\textwidth]{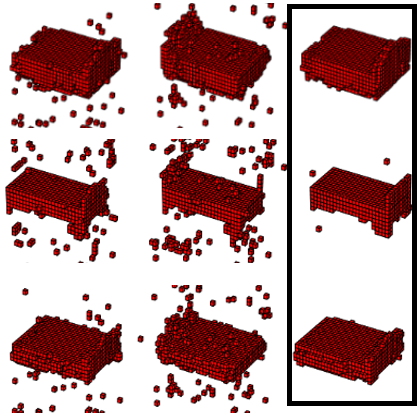}
  \caption{bed}
  \label{fig:sfig3d}
\end{subfigure}

\begin{subfigure}{.3\textwidth}
  \centering
  \includegraphics[width=\textwidth]{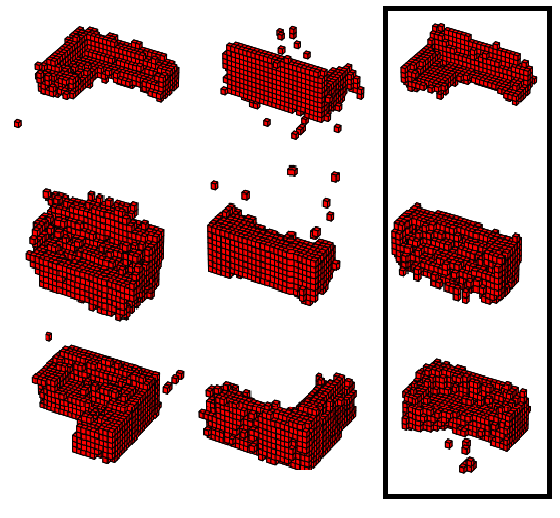}
  \caption{sofa}
 \label{fig:sfig3e}
\end{subfigure}%
\qquad
\begin{subfigure}{.3\textwidth}
  \centering
  \includegraphics[width=\textwidth]{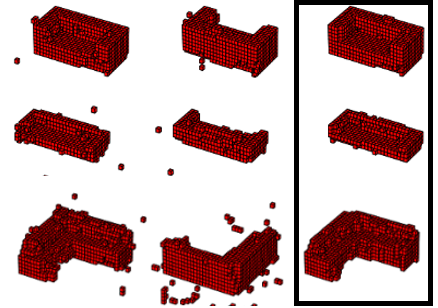}
  \caption{sofa}
  \label{fig:sfig3f}
\end{subfigure}
\caption{Results with 3 classes (chair, bed and sofa) using 2-conditions (rotations).  The first two samples are the generated pairs, merged results are shown in boxes. (a), (c) and (e) show the pairs generated with standard conditional GAN. It is clearly visible that the samples belong to different objects. Standard conditional GAN fails to generate the same object in different conditions (rotations) as expected and the merged results are noisy. (b), (d) and (f) show the pairs generated with the proposed method. The samples are very similar and the merged results (intersection of samples) support this observation. Merged results are also mostly noise-free and have more detail compared to standard conditional GAN.}
\label{fig:fig3}
\end{figure}

Visual results prove that, standard conditional GAN fails to generate 3D models with the same attributes in different rotations. In 2-conditional case, it generates a chair with \ang{0} orientation, and a completely different chair with \ang{180} orientation for the same input value. On the other hand, the proposed system can generate 3D models of the same object category with same attributes with \ang{0} and \ang{180} orientations. Also the result of merge operation is given to show the intersection of models. Since intersection of noise is mostly empty, merged model is also mostly noise-free. For these 3 classes, system is proven to generate pair models on different rotations.

For additional training of the proposed method, samples are generated by keeping the input vector the same and setting the condition value differently. Then the outputs are merged and fed into the discriminator. Only the generator is updated in this step. Experiments show that, also updating the discriminator in this step causes overtraining and makes the system unstable. Since this step is for enforcing the 
generator to generate the same sample in different conditions, training of the discriminator is not necessary.

\subsubsection{Merge method:} 
\begin{figure}
\centering
\begin{subfigure}{.4\textwidth}
  \centering
  \includegraphics[width=\textwidth]{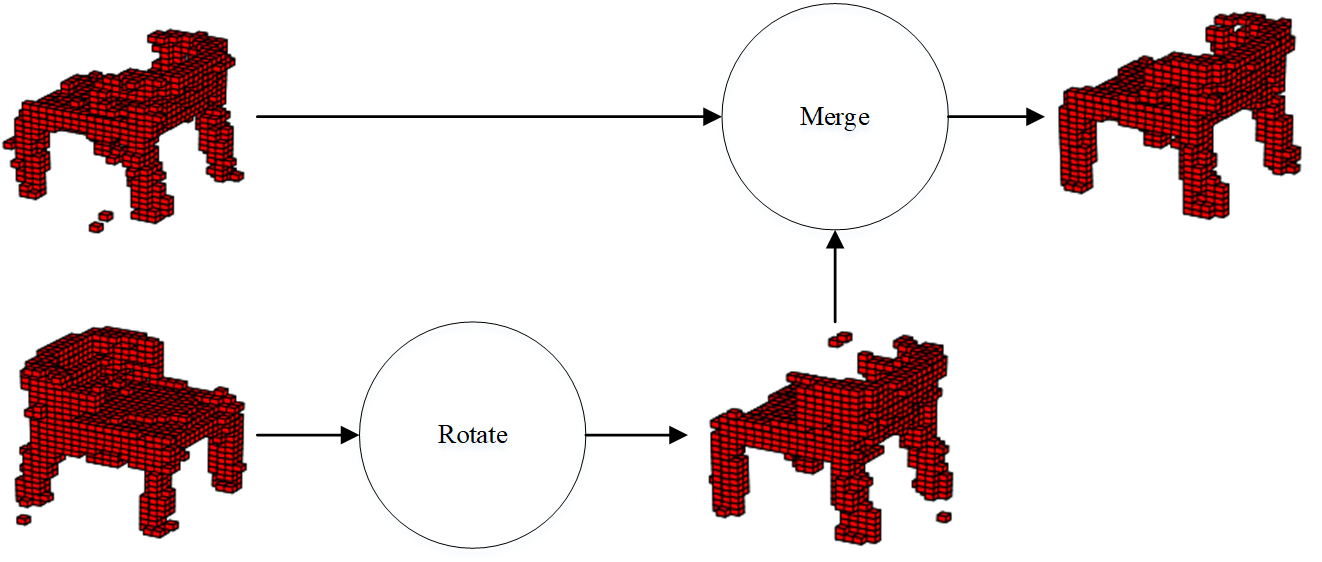}
  \caption{}
 \label{fig:sfig4a}
\end{subfigure}%
\qquad
\begin{subfigure}{.4\textwidth}
  \centering
  \includegraphics[width=\textwidth]{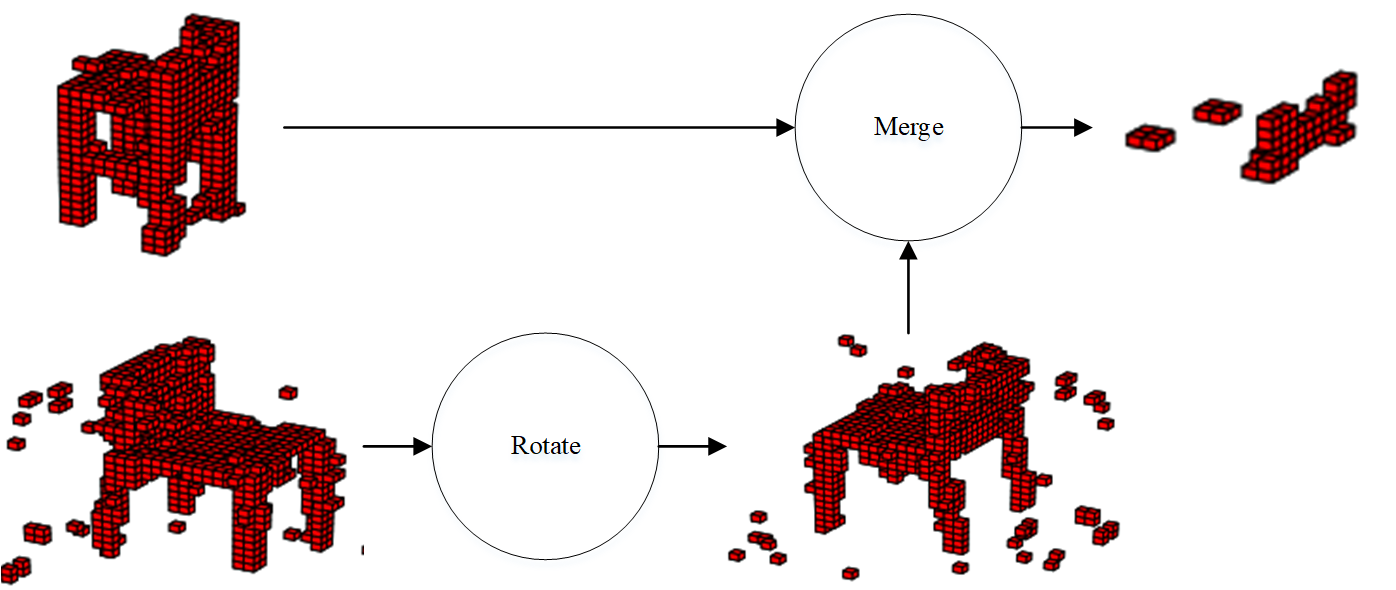}
  \caption{}
  \label{fig:sfig4b}
\end{subfigure}
\caption{Examples of merging operation. After generating pairs, one of the pairs is aligned with the other. Second sample is rotated to align with the first one in these examples. Then aligned samples are merged to form a new one. Simple averaging is applied to aligned pairs to get the intersection. (a) The result of the merging operation will be similar to the generated samples if the samples are similar, (b) the result will be meaningless if the samples are different.}
\label{fig:fig4}
\end{figure}

Merging the generated samples is domain specific. For our case, generated samples are 3D voxelized models with values between 0 and 1 representing the probability of the existence of the unit cube on that location. First aligning the samples generated with different orientations and then simply averaging their 3D matrices, we get the merged result. In Figure \ref{fig:fig4}, we illustrate the merging procedure with a 2-conditional case with chair dataset. Generator will output two chairs with \ang{0} and \ang{180} rotations respectively. We can simply rotate the second model \ang{180} to align both samples. Then, we average these 3D matrices. By averaging we get the probability of the existence of unit cubes in each location taking both outputs into account. If chairs are similar, the intersection of them will also be a similar chair (Fig. \ref{fig:fig4}(a)) and if the chairs are not similar, their intersection will be meaningless (Fig. \ref{fig:fig4}(b)). By feeding these merged results into the discriminator, we make the network evaluate the intersection model and train the generator using this information.

\subsubsection{Results:} The proposed framework has been implemented using Tensorflow \cite{tensorflow2015-whitepaper} version 1.4 and tested with 3 classes: chair, bed and sofa. The results are observed after training the model for 1500 epochs with the whole dataset. Dataset is divided into batches of 128 samples. For comparison, we used the conditional GAN that we adapted for 3D model generation as the baseline method. Both systems have been trained with the same parameters and same data. Results are generated with the same input and different condition values. To visualize the results, binary voxelization is used with a threshold of 0.5. Fig. \ref{fig:fig3} shows the visual results. Note that the presented results are visualizations of raw output without any post processing or noise reduction.

As there is no established metric for the evaluation of generated samples, we introduce 2 different evaluation metrics: Average Absolute Difference (AAD) and Average Voxel Agreement Ratio (AVAR). 

Raw outputs are 3D matrices for each generated model and each element of these matrices is a probability value between 0 and 1. For the calculation of AAD with n-conditions, first, the generated models $S_1,\dotsc, S_n$ aligned with $S_0$ to get $S^R_1,\dotsc, S^R_n$  then AAD can be formulated as follows:

\begin{equation}
AAD = \frac{\sum_{i=0}^{n-1}\frac{\sum_{\forall x,y,z}|S_{i}^{R}(x,y,z)-M(x,y,z)|}{total~\#~of~matrix~elements}}{n}
\end{equation}

As a result of AAD a single difference metric is obtained for that object. A lower AAD value indicates agreement of the generated models with the merged model and it is desired to have an AAD value closer to 0. 

For the calculation of Average Voxel Agreement Ratio (AVAR), first the aligned 3D matrices are binarized with a threshold of 0.5 to form voxelized  $S^{RB}_i$ $M^B$ and  then Average Voxel Agreement Ratio (AVAR) can be formulated as:

\begin{equation}
AVAR = \frac{\sum_{i=0}^{n-1}\frac{\sum_{\forall x,y,z}S_{i}^{RB}(x,y,z)\bigwedge M^{B}(x,y,z)}{\sum_{\forall x,y,z}S_{i}^{RB}(x,y,z)}}{n}
\end{equation}

where $\bigwedge $ is the binary logical AND operator. AVAR value of 0 indicates disagreement while a value of 1 indicates agreement of the models with the merged model and it is desired to have an AVAR value closer to 1.

Results for 2-conditions and a batch of 128 pairs are given in Table 1. AAD and AVAR results are calculated separately for each pair in the batch and then averaged to get a single result for the batch. The results show that the proposed method reduces the average difference significantly; 3, 2.4 and 4.5 times for chair, bed and sofa respectively. Here the results are highly dependent to object class. Different beds and sofas are naturally more similar than different chairs. While different bed shapes are mostly same except headboards, chairs can be very different considering stools, seats etc. Also we can see it in the results, the proposed method improved the similarity of generated chair pairs from 0.32 to 0.79. While the generated chair pairs are very different with the baseline method, the proposed method generated very similar pairs. For bed and sofa the baseline similarities are 0.69 and 0.74, relatively more similar as expected. The proposed method improved the results to 0.89 and 0.95 for bed and sofa respectively by converging to the same model. 
\begin{table}
\caption{Comparison of the proposed method with baseline using different object classes for 2-conditions and a batch (128) of pairs. AAD: Average Absolute Difference between generated matrices, AVAR: Average Voxel Agreement Ratio.}\label{tab1}
\centering

\normalsize
\begin{tabular}{|c|c|c|c|c|c|c|}
\hline
\multirow{2}{*}{} & \multicolumn{2}{c|}{Chair} & \multicolumn{2}{c|}{Bed} & \multicolumn{2}{c|}{Sofa} \\ \cline{2-7} 
                  &~ AAD~          & ~AVAR~        & ~AAD~         & ~AVAR~       & ~AAD~          & ~AVAR~       \\ \hhline{|-|=|=|=|=|=|=|}
Baseline          & 0.027        & 0.32        & 0.029       & 0.69       & 0.018        & 0.74       \\ \hline
Proposed          & 0.009        & 0.79        & 0.012       & 0.89       & 0.004        & 0.95       \\ \hline
\end{tabular}%

\end{table}

\begin{table}
\caption{Comparison of the proposed method with baseline using different object classes for 4-conditions. The same metrics are used as in the 2-condition case.}\label{tab2}
\centering
\normalsize
\begin{tabular}{|c|c|c|c|c|c|c|}
\hline
\multirow{2}{*}{} & \multicolumn{2}{c|}{Chair} & \multicolumn{2}{c|}{Bed} & \multicolumn{2}{c|}{Sofa} \\ \cline{2-7} 
                  &~ AAD~          & ~AVAR~        & ~AAD~         & ~AVAR~       & ~AAD~          & ~AVAR~       \\ \hhline{|-|=|=|=|=|=|=|}
Baseline          & 0.034        & 0.36        & 0.043       & 0.65       & 0.034        & 0.62       \\ \hline
Proposed          & 0.024        & 0.61        & 0.021       & 0.82       & 0.013        & 0.90       \\ \hline
\end{tabular}
\end{table}

The proposed system has also been tested with 4-conditions. For 4 orientations, condition 0, 1, 2 and 3 are used for \ang{0}, \ang{90}, \ang{180} and \ang{270} respectively. Also for merging operation, all generated samples are aligned with the first sample with \ang{0} rotation. For that purpose 2\textsuperscript{nd}, 3\textsuperscript{rd} and 4\textsuperscript{th} samples are rotated by \ang{270}, \ang{180} and \ang{90} respectively. After aligning all 4 samples, they are merged into a single model by averaging.

\begin{figure}

\centering
\begin{subfigure}{.45\textwidth}
  \centering
  \includegraphics[width=\textwidth]{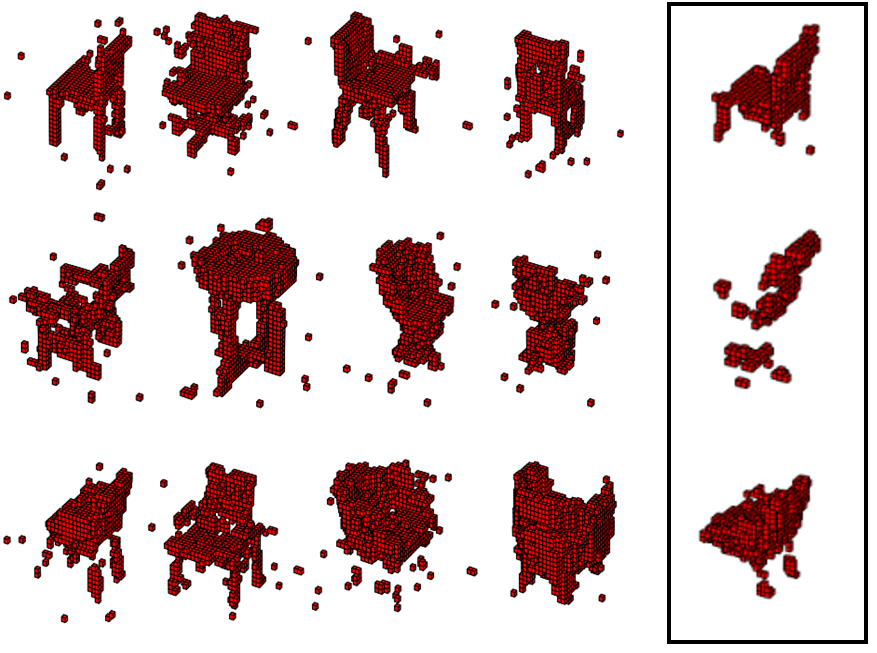}
  \caption{chair}
 \label{fig:sfig5a}
\end{subfigure}%
\qquad
\begin{subfigure}{.4\textwidth}
  \centering
  \includegraphics[width=\textwidth]{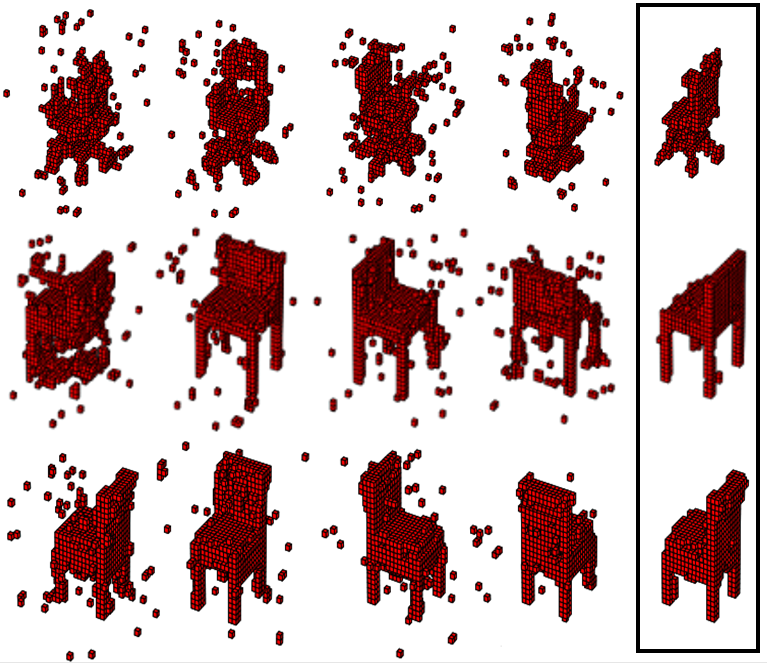}
  \caption{chair}
  \label{fig:sfig5b}
\end{subfigure}

\begin{subfigure}{.45\textwidth}
  \centering
  \includegraphics[width=\textwidth]{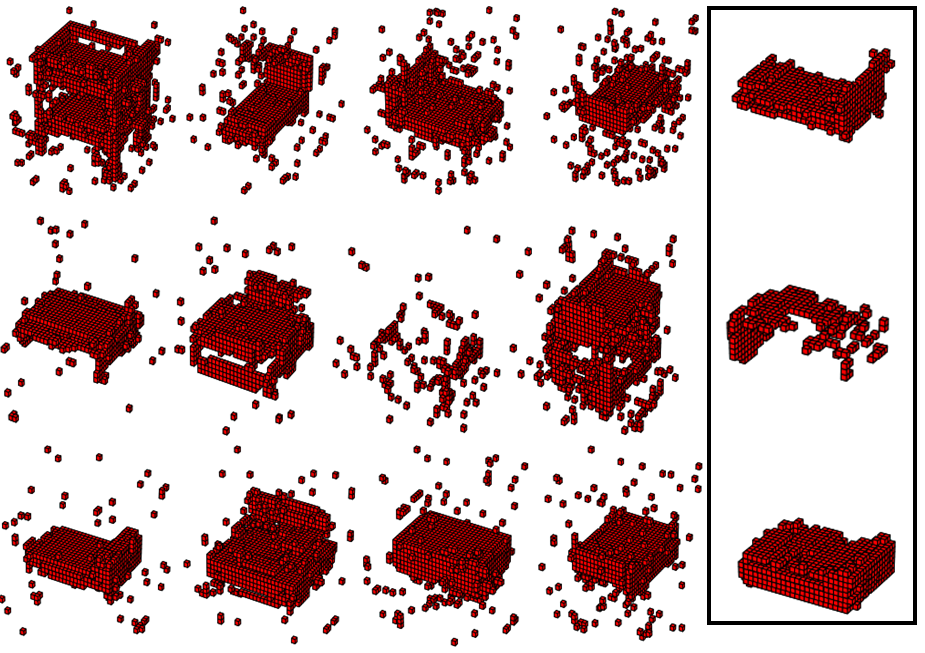}
  \caption{bed}
 \label{fig:sfig5c}
\end{subfigure}%
\qquad
\begin{subfigure}{.4\textwidth}
  \centering
  \includegraphics[width=\textwidth]{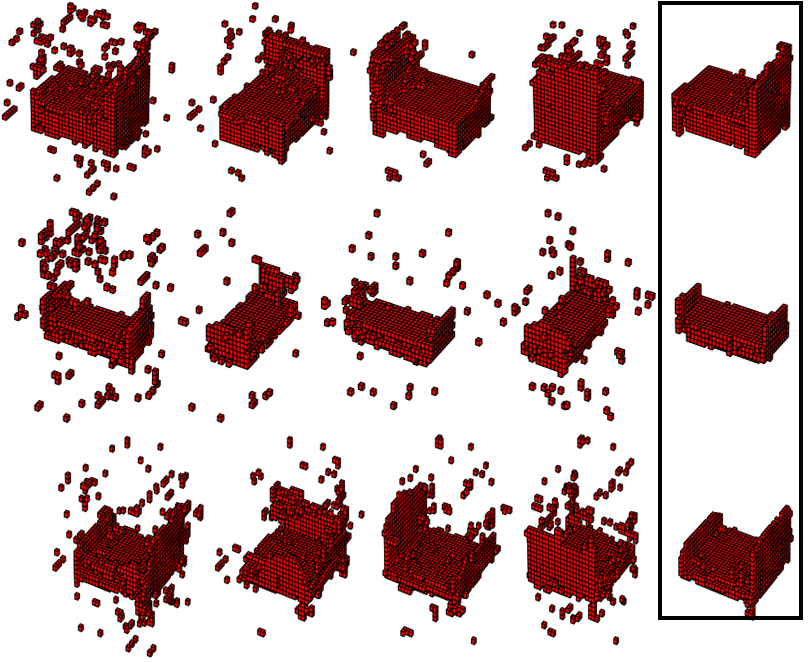}
  \caption{bed}
  \label{fig:sfig5d}
\end{subfigure}

\begin{subfigure}{.45\textwidth}
  \centering
  \includegraphics[width=\textwidth]{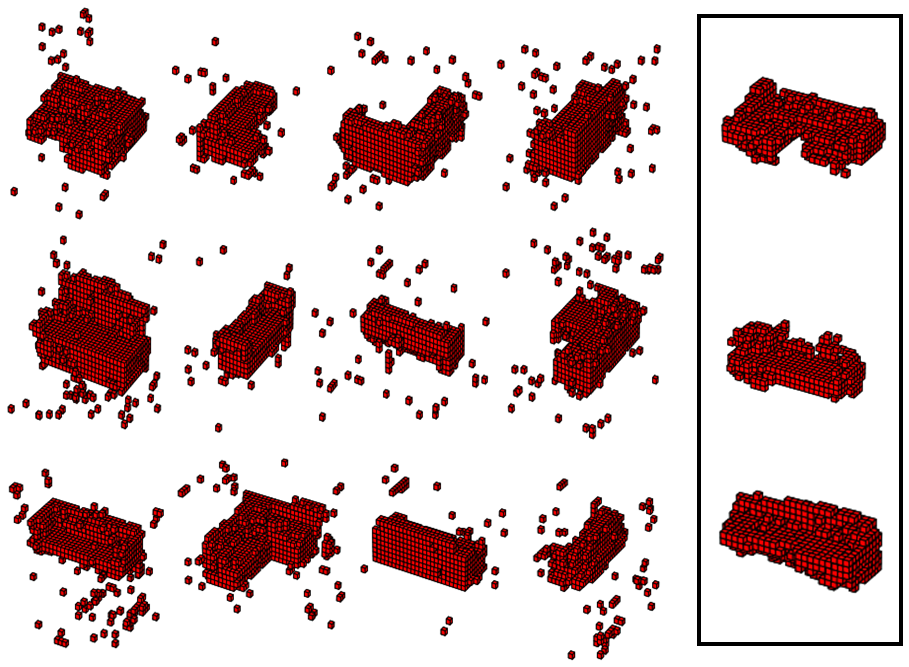}
  \caption{sofa}
 \label{fig:sfig5e}
\end{subfigure}%
\qquad
\begin{subfigure}{.4\textwidth}
  \centering
  \includegraphics[width=\textwidth]{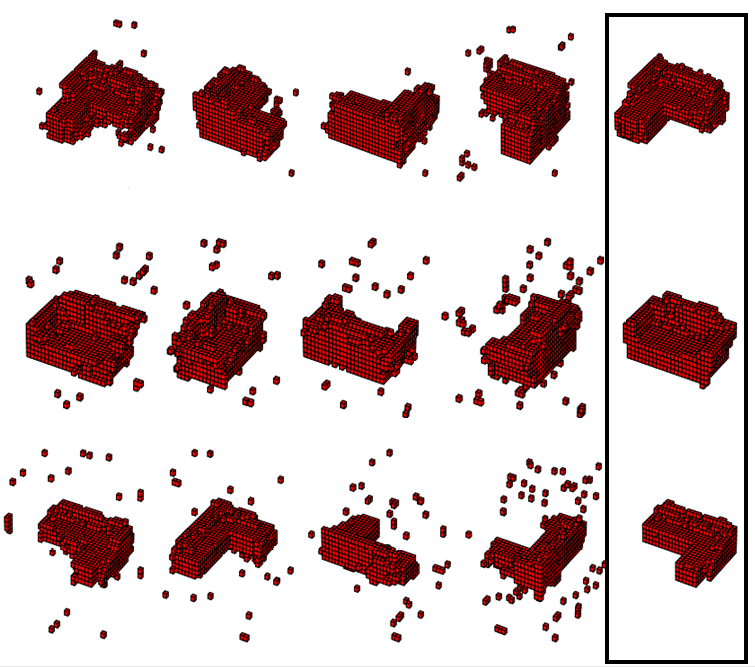}
  \caption{sofa}
  \label{fig:sfig5f}
\end{subfigure}
\caption{Visual results with 4 conditions. The first four samples are the generated objects, merged results are shown in boxes. (a), (c) and (e) show the objects and the merged result obtained with standard conditional GAN. (b), (d) and (f) show the objects and the merged result obtained with the proposed method The samples are very similar and the merged results (intersection of samples) support this claim. Merged results are also mostly noise-free and have more detail compared to standard conditional GAN.}
\label{fig:fig5}
\end{figure}

	Fig. \ref{fig:fig5} shows the visual results for 4-conditional case with the same experimental setup. Experimental results in terms of the same metrics are presented in Table \ref{tab2}. Standard conditional GAN generates 4 different chairs on 4 rotations. On the other hand the proposed method enforces the network to generate the same chair on 4 different rotations. Since the problem is more complex for 4 rotations, individual generated samples are noisier and have lower resolution. The improvement rates compared to the baseline are relatively lower than 2-condition case because of the increased complexity of the problem. To account for the increasing complexity of the model with higher number of conditions, more training data and/or higher number of epochs need to be used.While generating better samples with more training may seem crucial, it doesn’t change the behavior of the networks. Conditional GAN keeps generating different samples and proposed model generates paired samples with each training iteration.

\section{Conclusions and Future Work}
\label{sect:conc}

In this paper, we presented a new approach to generate paired 3D models with conditional GAN. First, we adapted the conditional GAN to generate 3D models on different rotations. Then, we integrated an additional training step to solve problem of generation of pair samples, which is a shortcoming of standard conditional GAN. The proposed method is generic and it can be integrated into any conditional GAN. The results show the potential of the proposed method for the popular problem of joint distribution learning in GANs.

We demonstrated that proposed method works successfully for 3D voxel models on 2 and 4 orientations. Visual results and the objective evaluation metrics confirm the success of the proposed method. The difference between generated models are reduced significantly in terms of the average difference. The merged samples create noise-free high-resolution instances of the objects. This approach can also be used for generating better samples compared to traditional GAN for a particular object class. 

The extension of the method to work with higher number of conditions is trivial. However, as the training of the system takes a long time, we leave the experiments with higher number of conditions and classes as a future work. The proposed solution is generic and could be applied to other types of data. As a next step, we are aiming to test the method on generation of 2D images to investigate the validity of the method for different data types.

\clearpage
%
%
%


\begin{thebibliography}{10}

\bibitem{tensorflow2015-whitepaper}
Abadi, M., Agarwal, A., Barham, P., Brevdo, E., Chen, Z., Citro, C., Corrado,
   G.S., Davis, A., Dean, J., Devin, M., Ghemawat, S., Goodfellow, I.,  
Harp, A.,
   Irving, G., Isard, M., Jia, Y., Jozefowicz, R., Kaiser, L., Kudlur, M.,
   Levenberg, J., Man\'{e}, D., Monga, R., Moore, S., Murray, D., Olah, C.,
   Schuster, M., Shlens, J., Steiner, B., Sutskever, I., Talwar, K.,  
Tucker, P.,
   Vanhoucke, V., Vasudevan, V., Vi\'{e}gas, F., Vinyals, O., Warden, P.,
   Wattenberg, M., Wicke, M., Yu, Y., Zheng, X.: {TensorFlow}: Large-scale
   machine learning on heterogeneous systems (2015),
   \url{https://www.tensorflow.org/}, software available from tensorflow.org

\bibitem{achlioptas2017representation}
Achlioptas, P., Diamanti, O., Mitliagkas, I., Guibas, L.: Representation
   learning and adversarial generation of 3d point clouds. arXiv preprint
   arXiv:1707.02392  (2017)

\bibitem{arjovsky2017wasserstein}
Arjovsky, M., Chintala, S., Bottou, L.: Wasserstein gan. arXiv preprint
   arXiv:1701.07875  (2017)

\bibitem{che2016mode}
Che, T., Li, Y., Jacob, A.P., Bengio, Y., Li, W.: Mode regularized generative
   adversarial networks. arXiv preprint arXiv:1612.02136  (2016)

\bibitem{chen2018syncgan}
Chen, W.C., Chen, C.W., Hu, M.C.: Syncgan: Synchronize the latent space of
   cross-modal generative adversarial networks. arXiv preprint arXiv:1804.00410
   (2018)

\bibitem{chen2016infogan}
Chen, X., Duan, Y., Houthooft, R., Schulman, J., Sutskever, I., Abbeel, P.:
   Infogan: Interpretable representation learning by information maximizing
   generative adversarial nets. In: Advances in neural information processing
   systems. pp. 2172--2180 (2016)

\bibitem{goodfellow2014generative}
Goodfellow, I., Pouget-Abadie, J., Mirza, M., Xu, B., Warde-Farley, D., Ozair,
   S., Courville, A., Bengio, Y.: Generative adversarial nets. In: Advances in
   neural information processing systems. pp. 2672--2680 (2014)

\bibitem{isola2017image}
Isola, P., Zhu, J.Y., Zhou, T., Efros, A.A.: Image-to-image translation with
   conditional adversarial networks. arXiv preprint arXiv:1611.07004 (2017)

\bibitem{jiang2017hierarchical}
Jiang, C., Marcus, P., et~al.: Hierarchical detail enhancing mesh-based shape
   generation with 3d generative adversarial network. arXiv preprint
   arXiv:1709.07581  (2017)

\bibitem{kim2017learning}
Kim, T., Cha, M., Kim, H., Lee, J.K., Kim, J.: Learning to discover
   cross-domain relations with generative adversarial networks. arXiv preprint
   arXiv:1703.05192  (2017)

\bibitem{kingma2015adam}
Kingma, D.P., Ba, J.L.: Adam: Amethod for stochastic optimization. In:
   Proceedings of the 3rd International Conference on Learning Representations
   (ICLR) (2015)

\bibitem{liu2016coupled}
Liu, M.Y., Tuzel, O.: Coupled generative adversarial networks. In: Advances in
   neural information processing systems. pp. 469--477 (2016)

\bibitem{luan2017deep}
Luan, F., Paris, S., Shechtman, E., Bala, K.: Deep photo style transfer. CoRR,
   abs/1703.07511  \textbf{2} (2017)

\bibitem{mao2017aligngan}
Mao, X., Li, Q., Xie, H.: Aligngan: Learning to align cross-domain images with
   conditional generative adversarial networks. arXiv preprint arXiv:1707.01400
   (2017)

\bibitem{mirza2014conditional}
Mirza, M., Osindero, S.: Conditional generative adversarial nets. arXiv
   preprint arXiv:1411.1784  (2014)

\bibitem{radford2015unsupervised}
Radford, A., Metz, L., Chintala, S.: Unsupervised representation learning with
   deep convolutional generative adversarial networks. arXiv preprint
   arXiv:1511.06434  (2015)

\bibitem{smith2017improved}
Smith, E., Meger, D.: Improved adversarial systems for 3d object  
generation and
   reconstruction. arXiv preprint arXiv:1707.09557  (2017)

\bibitem{wang2018perceptual}
Wang, C., Xu, C., Wang, C., Tao, D.: Perceptual adversarial networks for
   image-to-image transformation. IEEE Transactions on Image Processing
   \textbf{27}(8),  4066--4079 (2018)

\bibitem{wu2016learning}
Wu, J., Zhang, C., Xue, T., Freeman, B., Tenenbaum, J.: Learning a
   probabilistic latent space of object shapes via 3d generative-adversarial
   modeling. In: Advances in Neural Information Processing Systems. pp. 82--90
   (2016)

\bibitem{wu20153d}
Wu, Z., Song, S., Khosla, A., Yu, F., Zhang, L., Tang, X., Xiao, J.: 3d
   shapenets: A deep representation for volumetric shapes. In: Proceedings of
   the IEEE conference on computer vision and pattern recognition. pp.
   1912--1920 (2015)

\bibitem{zhu2017unpaired}
Zhu, J.Y., Park, T., Isola, P., Efros, A.A.: Unpaired image-to-image
   translation using cycle-consistent adversarial networks. arXiv preprint arXiv:1703.10593
   (2017)

\end{thebibliography}
\end{document}